\documentclass[a4paper,10pt,pdftex]{article}
\usepackage{fullpage}
\usepackage[utf8]{inputenc}
\usepackage{amsmath}
\usepackage{amssymb}
\usepackage{amsfonts}
\usepackage{amssymb}
\usepackage{algorithm, algpseudocode}
\usepackage[pdftex]{hyperref}
\usepackage{graphicx}
\usepackage{color}
\usepackage{bm}
\usepackage{multirow}
\usepackage{booktabs}
\usepackage{soul}
\usepackage{subfigure}
\usepackage[raggedright]{sidecap}
\sidecaptionvpos{figure}{c} % position of the text in sidecaption. c=center, t=top, b=bottom
\usepackage{multirow}
\usepackage[table]{xcolor}
\usepackage[square,sort,comma,numbers]{natbib}

\newcommand{\footremember}[2]{%
    \footnote{#2}
    \newcounter{#1}
    \setcounter{#1}{\value{footnote}}%
}

\newcommand{\footrecall}[1]{%
    \footnotemark[\value{#1}]%
} 

\begin{document}

\title{Learning representations of multivariate time series with missing data}

% \author[1]{Filippo Maria Bianchi}
% \author[3,4]{Lorenzo Livi}
% \author[1]{Karl {\O}yvind Mikalsen}
% \author[1]{Michael Kampffmeyer}
% \author[1,2]{Robert Jenssen}
% \address[1]{Machine Learning Group, UiT the Arctic University of Norway, http://site.uit.no/ml/}
% \address[2]{Norwegian Computing Center, Oslo, Norway}
% \address[3]{Departments of Computer Science and Mathematics, University of Manitoba, Winnipeg, MB R3T 2N2, Canada}
% \address[4]{Department of Computer Science, College of Engineering, Mathematics and Physical Sciences, University of Exeter, Exeter EX4 4QF, UK}

\author{%
  Filippo Maria Bianchi\footremember{corr}{Corresponding author: \texttt{filippombianchi@gmail.com}}\footremember{uit}{Machine Learning Group, UiT the Arctic University of Norway}%
  \and Lorenzo Livi \footremember{man}{Departments of Computer Science and Mathematics, University of Manitoba} \footremember{ext}{Department of Computer Science, College of Engineering, Mathematics and Physical Sciences, University of Exeter}%
  \and Karl {\O}yvind Mikalsen\footrecall{uit}%
  \and Michael Kampffmeyer\footrecall{uit}% \footnote{Germany?}
  \and Robert Jenssen \footrecall{uit} \footremember{nrs}{Norwegian Computing Center, Oslo, Norway}
}
\date{}

\maketitle

\begin{abstract} 
Learning compressed representations of multivariate time series (MTS) facilitates data analysis in the presence of noise and redundant information, and for a large number of variates and time steps.
However, classical dimensionality reduction approaches are designed for vectorial data and cannot deal explicitly with missing values.
In this work, we propose a novel autoencoder architecture based on recurrent neural networks to generate compressed representations of MTS.
The proposed model can process inputs characterized by variable lengths and it is specifically designed to handle missing data.
Our autoencoder learns fixed-length vectorial representations, whose pairwise similarities are aligned to a kernel function that operates in input space and that handles missing values.
This allows to learn \textit{good} representations, even in the presence of a significant amount of missing data.
To show the effectiveness of the proposed approach, we evaluate the quality of the learned representations in several classification tasks, including those involving medical data, and we compare to other methods for dimensionality reduction.
Successively, we design two frameworks based on the proposed architecture: one for imputing missing data and another for one-class classification.
Finally, we analyze under what circumstances an autoencoder with recurrent layers can learn better compressed representations of MTS than feed-forward architectures.
\end{abstract}

%%%%%%%%%%%%%%%%%%%%%%%%%%%%%%%%%%%%%%%%%%%%%%%%%%%
%%%%%%%%%%%%%%%%% INTRODUCTION   %%%%%%%%%%%%%%%%%%
%%%%%%%%%%%%%%%%%%%%%%%%%%%%%%%%%%%%%%%%%%%%%%%%%%%
\section{Introduction}

% MULTIVARIATE TIME SERIES DATA
Real-valued multivariate time series (MTS) allow to characterize the evolution of complex systems and is the core component in many research fields and application domains~\cite{chatfield2016analysis, 7765110}.
MTS analysis should account for relationships across variables and time steps, and, at the same time, deal with unequal time lengths~\cite{langkvist2014review, JAIN201935, OREGI2019506}.
Thanks to their ability to capture long-term dependencies, recurrent neural networks (RNNs) achieve state of the art results in tasks involving time series with one or more variables. 
In the latter case, RNNs can also be coupled with convolutional neural networks to explicitly model relationships across the different variables~\cite{song2018deep}.

In time series analysis it is important to handle missing values, which are commonly found in real-world data such as electronic health records (EHR)~\cite{che2017time, DING2012919}, and are usually filled with imputation techniques in a pre-processing phase~\cite{FARHANGFAR20083692}.
However, unless data are missing completely at random~\cite{heitjan1996distinguishing}, imputation destroys useful information contained in the missingness patterns.
Furthermore, an imputation method may introduce a strong bias that influences the outcome of the analysis, especially for large fractions of missing values~\cite{little2014statistical}.
A data-driven approach has recently been proposed to learn when to switch between two particular types of imputation~\cite{Che2018}, but it relies on strong assumptions that are suitable only for specific applications.

% IMPORTANCE OF LEARNING REPRESENTATIONS
A proper representation of the data is crucial in machine learning applications~\cite{LI20153542}.
While traditional approaches rely on manual feature engineering that requires time and domain expertise, representation learning aims at automatically producing features describing the original data~\cite{bengio2013representation}. 
Dimensionality reduction has been a fundamental research topic in machine learning~\cite{jenssen2010kernel, harandi2017dimensionality, MIKALSEN2019257} and its application in representation learning for extracting relevant information and generating compressed representations facilitates analysis and processing of data~\cite{langkvist2014review, WANG201955, TUNCEL2018202}.
This is particularly important in the case of MTS, which often contain noise and redundant information with a large number of variables and time steps~\cite{TUNCEL2018202}.
However, classical dimensionality reduction approaches are not designed to process sequential data, especially in the presence of missing values and variable input size.

\subsection{Contributions}
In this paper, we propose a novel neural network architecture, called the \textit{Temporal Kernelized Autoencoder} (TKAE), to learn compressed representations of real-valued MTS that contain missing data and that may be characterized by unequal lengths.
We assume data to be missing at random (MAR), i.e., the probability of data to be missing depends on the observed variables and the missingness patterns might be useful to characterize the data.
Under the MAR assumption, once one has conditioned on all the data at hand, any remaining missingness is completely random (i.e., it does not depend on some unobserved variable). 
The MAR assumption covers many practical cases and includes also those situations where data are missing completely at random, i.e., the probability of missing data is independent from any other variable and there are no missingness patterns. 

Our contributions are summarized in the following.
\newline

\textbf{Learning compressed representations of MTS with missing data.}
% KERNEL ALIGNMENT
To handle missing data effectively, and, at the same time, avoid the undesired biases introduced by imputation, we propose a \textit{kernel alignment} procedure~\cite{kernelAE} that matches the dot product matrix of the learned representations with a kernel matrix. 
Specifically, we exploit the recently-proposed Time series Cluster Kernel (TCK)~\cite{mikalsen2017time}, which computes similarities between MTS with missing values without using imputation. By doing so, we generate representations that preserve unbiased pairwise similarities between MTS even in the presence of large amounts of missing data. 
\\
% VAR LENGTH
The encoder and the decoder in the TKAE are implemented by stacking multiple RNNs, allowing to generate a \emph{fixed-size} vectorial representation of input MTS with variable-lengths.
To better capture time dependencies, we use a bidirectional RNN~\cite{6638947} in the encoder.
The final states of the forward and backward RNNs are combined by a dense nonlinear layer that reduces the dimensionality of the representation.
\\
% PROPERTIES
The proposed architecture serves different purposes.
First of all, it transports the data from a complex input domain to a low-dimensional vector space while preserving the original relationships between inputs described by the TCK.
Once represented as ordinary vectors, the MTS can then be processed by standard classifiers or by unsupervised machine learning algorithms~\cite{Xing:2010:BSS:1882471.1882478}, and their indexing and retrieval are more efficient \cite{7179089, chung2016unsupervised}.
Furthermore, when the dimensionality of the data is reduced, the models can potentially be trained with fewer samples.
\newline

% APPLICATIONS

\textbf{Frameworks for missing data imputation and anomaly detection.}
Contrarily to other nonlinear dimensionality reduction techniques, the TKAE provides a decoder that yields an explicit mapping back to the input space. 
We exploit this feature to implement frameworks for two different applications.
Specifically, we use the TKAE and its decoder to (i) implement an imputation method that leverages the generalization capability of the decoder reconstruction, rather than relying on \textit{a-priori} assumptions that may introduce stronger biases, and (ii) design an anomaly detector based on the reconstruction error of the inputs.
\newline

% ANALYSIS
\textbf{Analysis of how RNNs encode MTS.}
We provide a detailed analysis of the effect of using RNNs for encoding MTS, which is the mechanism adopted in TKAE to handle inputs with variable lengths.
Despite the popularity of AEs based on RNNs~\cite{sutskever2014sequence} for applications focused on text~\cite{DBLP:journals/corr/BowmanVVDJB15}, speech~\cite{SU2017397}, and video data~\cite{srivastava2015unsupervised}, significantly fewer research efforts have been devoted so far in applying these architectures to real-valued MTS and, in general, in the context of dynamical systems.
To fill this gap, we investigate under which circumstances recurrent architectures generate better compressed representations of MTS than feed-forward networks, which use padding to obtain inputs of equal length.
Results show that the effectiveness of the RNNs is related to specific intrinsic properties of the MTS.

% PAPER ORGANIZATION
\subsection{Paper organization}
The paper is organized as follows.
In Sec.~\ref{sec:methods} we first provide the background for existing AE models and the TCK. 
Then, we introduce the proposed TKAE architecture.
In Sec.~\ref{sec:experiments}, we evaluate the TKAE's capability to learn good representations both on controlled tests and real-world MTS classification datasets.
Results confirm that our method significantly improves the quality of the representations as the percentage of missing data increases.
Successively, in~\ref{sec:case_stud} we propose two different frameworks that exploit the properties of the TKAE \emph{decoder} for (i) imputing missing data and (ii) building a one-class classifier.
We achieve competitive results for the imputation task, while we outperform other state-of-the-art methods in one-class classification.
In Sec.~\ref{sec:analysis}, we perform an in-depth analysis to investigate which MTS are better represented by an AE with RNNs.
We report several results obtained in controlled environments, as well as on benchmark data, to support the findings in our analysis.
Finally, Sec.~\ref{sec:conclusion} reports our conclusions.

%%%%%%%%%%%%%%%%%%%%%%%%%%%%%%%%%%%%%%%%%%%%
%%%%%%%%%%%%%%%%% METHODS %%%%%%%%%%%%%%%%%%
%%%%%%%%%%%%%%%%%%%%%%%%%%%%%%%%%%%%%%%%%%%%
\section{Methods}
\label{sec:methods}

% --------------- AUTOENCODER --------------- 
\subsection{The Autoencoder}
\label{sec:ae}

The AE is a neural network traditionally conceived as a non-linear dimensionality reduction algorithm~\cite{Hinton504}, which has been further exploited to learn representations in deep architectures~\cite{bengio2009learning} and to pre-train neural network layers~\cite{erhan2010does}.
An AE simultaneously learns two functions; the first one, called the \textit{encoder}, is a mapping from an input domain, $\mathbb{R}^{D_x}$, to a hidden representation (\textit{code}) in $\mathbb{R}^{D_z}$.
The second function, the \textit{decoder}, maps from $\mathbb{R}^{D_z}$ back to $\mathbb{R}^{D_x}$.
The encoding function $\phi: \mathbb{R}^{D_x} \rightarrow \mathbb{R}^{D_z}$ and the decoding function $\psi: \mathbb{R}^{D_z} \rightarrow \mathbb{R}^{D_x}$ are defined as
\begin{equation}
\begin{aligned}
\label{eq:encoding_decoding}
\mathbf{z} = \phi(\mathbf{x}; \boldsymbol{\theta}_E); \;
\mathbf{\tilde{x}} = \psi(\mathbf{z}; \boldsymbol{\theta}_D),
\end{aligned}
\end{equation}
where $\mathbf{x}\in\mathbb{R}^{D_x}$, $\mathbf{z}\in\mathbb{R}^{D_z}$, and $\mathbf{\tilde{x}}\in\mathbb{R}^{D_x}$ denote a sample from the input space, its hidden representation, and its reconstruction given by the decoder, respectively.
The encoder $\phi(\cdot)$ is usually implemented by stacking dense layers of neurons equipped with nonlinear activation functions.
The decoder $\psi(\cdot)$ is architecturally symmetric to the encoder and operates in reverse direction; when inputs are real-valued vectors, the decoder's squashing nonlinearities are often replaced by linear activations \cite{vincent2010stacked}.
Finally, $\boldsymbol{\theta}_E$ and $\boldsymbol{\theta}_D$ are the trainable parameters of the encoder and decoder, respectively. 
The parameters are the connection weights and biases of each layer $m$, i.e., $\boldsymbol{\theta}_E = \{\mathbf{W}_{E}^{(m)}, \mathbf{b}_{E}^{(m)} \}$ and $\boldsymbol{\theta}_D = \{\mathbf{W}_{D}^{(m)}, \mathbf{b}_{D}^{(m)} \}$.
AEs are trained to minimize the discrepancy between the input $\mathbf{x}$ and its reconstruction $\mathbf{\tilde{x}}$.
In the case of real-valued inputs, this is usually achieved by minimizing a loss $L_r$ implemented as the empirical Mean Squared Error (MSE).
It has been shown that for real-valued data, when the MSE between original and reconstructed input is minimized, the learned representations are \emph{good} in the sense that the amount of mutual information with respect to the input is maximized~\cite{vincent2010stacked}.

%\begin{equation}
%    \label{eq:distortion}
%    L_r = \mathrm{MSE}(\mathbf{x}, \mathbf{\tilde{x}}) = \frac{1}{N} \sum_{i=1}^{N} \| \mathbf{x}_i - \mathbf{\tilde{x}}_i %\|^{2}.
%\end{equation}

In this paper, we focus on AEs with a ``bottleneck'', which learn an under-complete representation of the input, i.e., $D_z < D_x$, retaining as much useful information as possible to allow an accurate reconstruction~\cite{Hinton504}.
The learned lossy, compressed representation of the input, can be exploited, e.g., for clustering and visualization tasks~\cite{makhzani2015adversarial}, or to train a classifier~\cite{ng2016dual}.
The bottleneck already provides a strong regularization as it limits the variance of the model. However, further regularization can be introduced by tying the encoder and decoder weights ($\mathbf{W}_{D} = \mathbf{W}^T_{E}$) or by adding a $\ell_2$-norm penalty to the loss function
\begin{equation}
    \label{eq:loss}
    L = L_r + \lambda L_2 = \mathrm{MSE}(\mathbf{x}, \mathbf{\tilde{x}}) + \lambda \lVert \mathbf{W} \rVert_{2}^{2},
\end{equation}
where $L_2$ is the $\ell_2$-norm of all model weights $\mathbf{W} = \{ \mathbf{W}_{D}, \mathbf{W}_{E} \}$ and $\lambda$ is the hyperparameter controlling the contribution of the regularization term.
%Further regularization schemes include noise injection in the input (Denoising AE \cite{vincent2010stacked}) and sparsity enforcement in the code by minimizing $l_1$ norm or divergence with respect to a given prior distribution \cite{gregor2010learning}.

% RNN & SEQ2SEQ
Recurrent neural networks (RNNs) are models excelling in capturing temporal dependencies in sequences~\cite{elman1990finding, bianchi2017recurrent} and are at the core of \textit{seq2seq} models~\cite{sutskever2014sequence}.
The latter, learns fixed-size representations of sequences with unequal length and, at the same time, generates variable-length outputs.

% attention
Modern \textit{seq2seq} architectures implement a powerful mechanism called \textit{attention}, which provides an inductive bias that facilitates the modeling of long-term dependencies and grants a more accurate decoding if the lengths of the input sequences varies considerably~\cite{BahdanauCB14, qin2017dual}.
However, models with attention provide a representation that is neither compact nor of fixed size and, henceforth, are not suitable for our purposes.
If fact, rather than learning a single vector representation for the whole input sequence, a model with attention maintains all the encoder states generated over time, which are combined by a time-varying \textit{decoding vector} at each decoding step.

% --------------- TCK --------------- 
\subsection{The Time Series Cluster Kernel}
\label{sec:tck}
The time series cluster kernel (TCK)~\cite{mikalsen2017time} is an algorithmic procedure to compute unsupervised kernel similarities among MTS containing missing data.
The TCK is able to model time series with missing data, under the MAR assumption.
The method is grounded on an ensemble learning approach that guarantees robustness with respect to hyperparameters.
This ensures that the TCK works well in unsupervised settings (the ones in which AEs actually operate), where it is not possible to tune hyperparameters by means of supervised cross-validation. 
The base models in the ensemble are Gaussian mixture models (GMMs), whose components are fit to the dataset.
By fitting GMMs with different numbers of mixtures, the TCK procedure generates partitions at different resolutions that capture both local and global structures in the data.

To further enhance diversity in the ensemble, each partition is evaluated on a random subset of MTS samples, attributes (variates), and time segments, using random initializations and randomly chosen hyperparameters. 
This also contributes to provide robustness with respect to hyperparameters, such as the number of mixture components.
To avoid imputation, missing data are analytically marginalized away in the likelihoods. 
To obtain the GMM posteriors, the likelihoods are multiplied with smooth priors, whose contribution becomes stronger as the percentage of missingness increases.
The TCK is then built by summing up, for each partition, the inner products between pairs of posterior assignments corresponding to different MTS.
More details on the TCK are provided in \ref{app:tck}.

% --------------- TKAE --------------- 
\subsection{The Temporal Kernelized Autoencoder}
\label{sec:tkae}

\begin{figure*}[th!]
    \centering
    \includegraphics[keepaspectratio,width=0.9\textwidth]{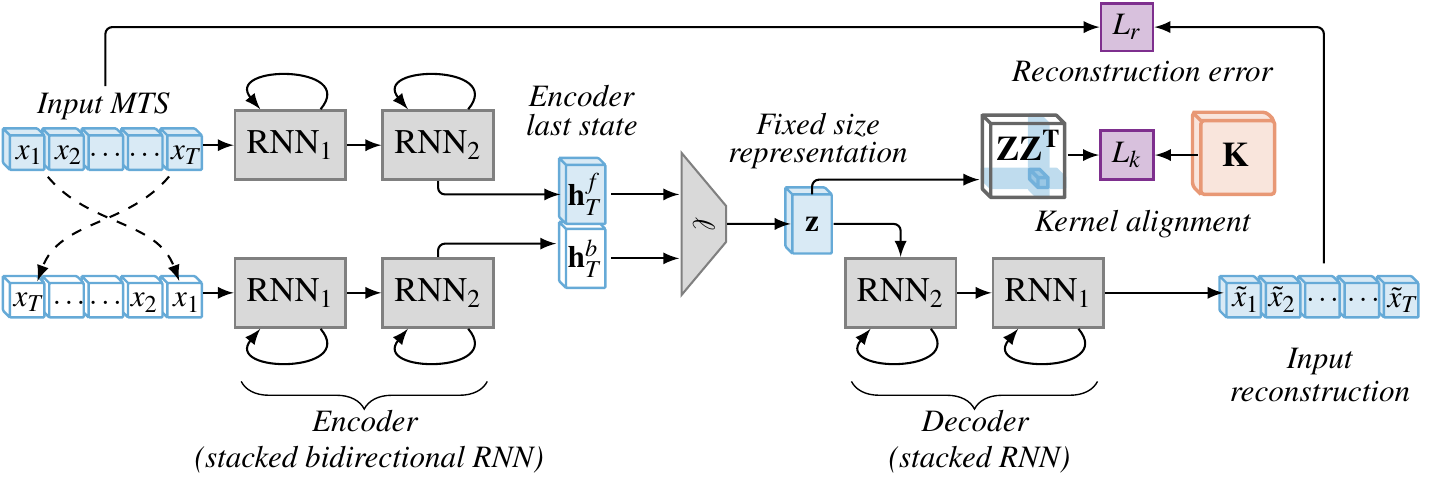}
    \caption{Schematic representation of TKAE. 
    Inputs are processed by a stacked bidirectional RNN. The last states obtained in forward $\mathbf{h}_T^f$ and backward $\mathbf{h}_T^b$ directions are combined by a dense layer $\ell$ to produce a fixed-size representation $\mathbf{z}$ of the input. $\mathbf{z}$ is used to initialize the state in the decoder, which is a stacked RNN operating in generative mode and is trained to reproduce inputs by minimizing the reconstruction error $\mathbf{L_r}$. 
    TKAE allows learning similarity-preserving representations of the inputs. 
    In particular, the matrix $\mathbf{Z}\mathbf{Z}^T$ containing the dot products of the representations of the MTS in the dataset is aligned, by means of a cost term $L_k$, to the kernel matrix $\mathbf{K}$. 
    The kernel matrix $\mathbf{K}$ is provided by the user as prior information to condition the representations.
    In our case, the kernel alignment generates representations whose relationships account for missing data in the input.
    }
\label{fig:TKAE_schema}
\end{figure*}

The \textit{Temporal Kernelized Autoencoder} (TKAE) is our proposed AE architecture, which is specifically conceived to learn compressed representations of variable-length MTS that may contain missing values. A schematic depiction of the TKAE is provided in Fig. \ref{fig:TKAE_schema}.

We assume that each MTS can be represented by a matrix $\mathbf{X}\in\mathbb{R}^{V \times T}$, where $V$ denotes the number of variates and $T$ is the number of time steps that may vary in each MTS.
Analogously to \textit{seq2seq}~\cite{sutskever2014sequence}, in TKAE the dense layers of standard AEs are replaced by RNNs, which process inputs sequentially and update their hidden state at each time step $t$ according to the following mapping,
\begin{equation}
    \label{eq:state_update}
    \mathbf{h}_t = \phi(\mathbf{x}_t,\mathbf{h}_{t-1}, \boldsymbol{\theta}_E),
\end{equation}
where $\boldsymbol{\theta}_E$ is the set of trainable parameters of the recurrent units.
The recurrent layers are composed of either gated recurrent units (GRU)~\cite{cho2014learning} or long short-term memory (LSTM)~\cite{hochreiter1997long} cells.
The choice of the cell is usually guided by the task at hand~\cite{chung2014empirical}.

% encoder
Conventional RNNs make use of previous inputs to build their current hidden representation~\cite{6638947}. 
However, in applications like MTS classification where the whole input is available at once, it is also possible to exploit the information contained in future inputs to generate the current state of the network.
For this reason, the encoder is implemented as a stacked bidirectional RNN~\cite{schuster1997bidirectional} consisting of two RNNs working in parallel, each one with $M$ layers of $D_z$ cells and transition function~\eqref{eq:state_update}. 
One of the RNNs captures input dependencies going \textit{backward} in time, whereas the other processes the same input but in reverse order, thus modeling relationships going \textit{forward} in time.
After the whole input is processed, the final states of the forward and backward RNN are denoted as $\mathbf{h}_T^f$ and $\mathbf{h}_T^b$, respectively.
While $\mathbf{h}_T^f$ is influenced by past observations, $\mathbf{h}_T^b$ depends on future ones. Hence, their combination can capture a wider range of temporal dependencies in the input.
In the TKAE, dense nonlinear layer $\ell$ combines the two states $\mathbf{h}_T^f$ and $\mathbf{h}_T^b$ and produces an output vector $\mathbf{z} \in \mathbb{R}^{D_z}$.
The latter, is the fixed-size, vectorial representation of the MTS.

% decoder
The decoder operates according to the following map,
\begin{equation}
    \label{eq:dec_out}
    \tilde{\mathbf{x}}_t = \psi(\mathbf{h}_t, \tilde{\mathbf{x}}_{t-1}, \boldsymbol{\theta}_D),
\end{equation}
where $\psi(\cdot, \cdot)$ is a stacked RNN with $M$ layers parametrized by $\boldsymbol{\theta}_D$ that operates in a generative mode, processing the previously generated output as a new input.
To initialize the decoder, we let its initial state $\mathbf{h}_0 = \mathbf{z}$ and first input $\tilde{\mathbf{x}}_0=\mathbf{0}$, which corresponds to an ``average input'' if the MTS are standardized.
The decoder iteratively produces outputs for $T$ steps, $T$ being the length of the input MTS.
Unequal lengths are naturally handled since the whole architecture is independent of $T$.

% training
The TKAE is trained end-to-end by means of stochastic gradient descent with scheduled sampling~\cite{NIPS2015_5956}.
More specifically, during training the decoder input at time $t$ is, with probability $p_s$, the decoder output at time $t-1$ (inference mode) and with probability $1-p_s$ the desired output at time $t-1$ (teacher forcing). 
Since the desired output is not available during the test phase, the decoder generates test data operating only in generative mode ($p_s=1$).
In most of our experiments, scheduled sampling improved the training convergence speed, providing a practical motivation for our choice.

% kernel alignment
Analogously to standard AEs, RNNs cannot directly process data with missing values, which are thus filled beforehand with some imputed value (0, mean value, last observed value)~\cite{schafer2002missing}.
However, imputation injects biases in the data that may negatively affect the quality of the representations and conceal potentially useful information contained in the missingness patterns.
To overcome these shortcomings, we introduce a \textit{kernel alignment} procedure~\cite{kernelAE} that allows us to preserve the pairwise similarities of the inputs in the learned representations. 
These pairwise similarities are encoded in a positive semi-definite matrix $\mathbf{K}$ that is defined by the designer and passed as input to the model.
In our case, by choosing the TCK matrix as $\mathbf{K}$, the learned representations will also account for missing data.

Kernel alignment is implemented by an additional regularization term in the loss function \eqref{eq:loss}, which becomes
\begin{equation}
    \label{eq:TKAE_loss}
    L = L_r + \lambda L_2 + \alpha L_k.
\end{equation}
$L_k$ is the kernel alignment cost, which takes the form of a normalized Frobenius norm of the difference between two matrices: $\mathbf{K}$ and $\mathbf{ZZ}^T$, the dot product matrix between the hidden representations $\mathbf{z}$ of the input MTS.
More specifically, the $L_k$ term is defined as
\begin{equation}
\label{eq:regularization}
    L_k = \Bigg{\lVert} \frac{\mathbf{ZZ}^T}{\|\mathbf{ZZ}^T\|_F} - \frac{\mathbf{K}}{\|\mathbf{K}\|_F} \Bigg{\rVert}_{F},
\end{equation}
where $\mathbf{Z} \in \mathbb{R}^{N \times D_z}$ is a matrix of hidden representations relative to the $N$ MTS in the dataset (or, more specifically, in the current mini-batch).
Finally, $\alpha, \lambda \geq 0$ are hyperparameters controlling the contribution of alignment and regularization costs in the overall loss function.

%%%%%%%%%%%%%%%%%%%%%%%%%%%%%%%%%%%%%%%%%%%%%%%
%%%%%%%%%%%%%%%%% EXPERIMENTS %%%%%%%%%%%%%%%%%
%%%%%%%%%%%%%%%%%%%%%%%%%%%%%%%%%%%%%%%%%%%%%%%
\section{Experiments}
\label{sec:experiments}

% summary of the experiments
The experimental section is organized as follows.

\begin{enumerate}
    \item \textbf{Quantitative evaluations of the representations in the presence of missing data.} In Sec.~\ref{sec:missing_data}, we evaluate the effectiveness of the kernel alignment for generating compressed representations in the presence of missing data by computing how accurately the representations are classified.
    Results show that the kernel alignment with the TCK greatly improves the classification accuracy of the MTS representations in the presence of large amounts of missing data.
    \item \textbf{Design and evaluation of decoder-based frameworks.} In Sec.~\ref{sec:case_stud}, we propose two novel frameworks based on the TKAE decoder for (i) imputing missing data and (ii) for one-class classification. 
    Both frameworks exploit not only the TKAE hidden representation but also its decoder, as the results are computed by mapping the compressed representations back to the input space.
    The proposed frameworks are tested on two different case-studies and compared with other methods.
\end{enumerate}

% experimental setup 
\paragraph{\textbf{Experimental setup}}
In the following, we compare the TKAE with methods for dimensionality reduction: PCA, a standard AE, and other RNN-based architectures. The learned compressed representations have the same dimensionality for all models that are taken into account.
Let $D_x$ be the input dimensionality; in TKAE $D_x = V$, as it processes recursively each single time step. 
On the other hand, the MTS must be unfolded into vectors when processed by PCA and AE.
Therefore, in AE and PCA the input dimensionality is $D_x = V \cdot T$.
We let $D_z$ be the size of the compressed representations, which corresponds to the number of RNN cells in each TKAE layer, the size of the innermost layer in AE, and the number of principal components in PCA, respectively.
In all experiments we use an AE with 3 hidden layers, $\{D_x, 30, D_z, 30, D_x\}$; the number of neurons in the intermediate layers ($30$) has been set after preliminary experiments and is not a critical hyperparameter (comparable results were obtained using $20$ or $40$ neurons).
As a measure of performance, we consider the MSE between the original test data and their reconstruction as produced by each model.
In each experiment, we train the models for 5000 epochs with mini-batches containing 32 MTS using the Adam optimizer~\cite{kingma2014adam} with an initial learning rate of $0.001$.
We independently standardize each variate of the MTS in all datasets.
In each experiment, and for each method, we identify the optimal hyperparameters with $k$-fold cross-validation evaluated on the reconstruction error (or, in general, on the unsupervised loss function) and we report the average results on the test set, obtained in $10$ independent runs.
We consider only TKAE models with a maximum of three hidden layers of either LSTM or GRU cells, as deeper models generally improve performance slightly at the cost of greater complexity~\cite{reimers2017optimal}.
When kernel alignment is not used ($\alpha=0$), we refer to the TKAE simply as the TAE.

% datasets
\paragraph{\textbf{Datasets}}
We consider several real-world dataset from the UCI and UCR~\footnote{\url{archive.ics.uci.edu/ml/datasets.html}, \url{www.cs.ucr.edu/~eamonn/time_series_data}} repositories and two medical datasets; details are reported in Tab.~\ref{tab:dataset_details}.
The datasets have been selected in order to cover a wide variety of cases, in terms of training/test sets size, number of variates $V$, number of classes and (variable) lengths $T$ of the MTS.

% :::::::::::::::::::::::::: TAB DATASET DESCRIPTION ::::::::::::::::::::::::::
\bgroup
\def\arraystretch{0.95} %vertical padding
\setlength\tabcolsep{.3em} %horizontal padding
\begin{table}[!ht]
\small
\centering
\caption{Benchmark time series datasets. Column 2 to 5 report the number of attributes, samples in training and test set, and classes, respectively. 
$T_{min}$ is the length of the shortest MTS in the dataset and $T_{max}$ the longest MTS.}
\label{tab:dataset_details}
\begin{tabular}{@{}ll@{ }@{ }l@{ }@{ }@{ }l@{ }@{ }@{ }ll@{ }l@{ }lc@{}}
\cmidrule[1.5pt]{1-8}
\textbf{Dataset} & $\boldsymbol{V}$ & \textbf{Train} & \textbf{Test} & \textbf{Classes} & $\boldsymbol{T}_{min}$ & $\boldsymbol{T}_{max}$ & \textbf{Source} \\
\cmidrule[.5pt]{1-8}
ECG & 1 & 500 & 4500 & 5 & 140 & 140 & UCR \\
%ECG2 & 2 & 100 & 100 & 2 & 39 & 152 & UCR \\
Libras & 2 & 180 & 180 & 15 & 45 & 45 & \cite{Baydogan2016} \\
%Char.Traj. & 3 & 300 & 2558 & 20 & 109 & 205 & UCI \\
Wafer & 6 & 298 & 896 & 2 & 104 & 198 & UCR \\
Jp. Vow. & 12 & 270 & 370 & 9 & 7 & 29 & UCI \\
Arab. Dig. & 13 & 6600 & 2200 & 10 & 4 & 93 & UCI \\
Auslan & 22 & 1140 & 1425 & 95 & 45 & 136 & UCI \\
EHR & 10 & 892 & 223 & 2 & 20 & 20 & \cite{2018arXiv180307879O} \\
Physionet & 2 & 8524 & 298 & 4 & 5 & 176 & \cite{clifford2017af} \\
\cmidrule[1.5pt]{1-8}
\end{tabular}
\end{table}
\egroup
% :::::::::::::::::::::::::::::::::::::::::::::::::::::::::::::::::::::::::::

The first medical dataset is the \textit{EHR} dataset, which contains blood samples collected over time, extracted from the Electronic Health Records of patients undergoing a gastrointestinal surgery at the University Hospital of North Norway in 2004--2012~\cite{2018arXiv180307879O}. 
Each patient is represented by a MTS of $V=10$ blood sample measurements collected for $T=20$ days after surgery.
We consider the problem of classifying patients with and without surgical site infections from their blood samples.
The dataset consists of $N=883$ MTS, of which $232$ pertain to infected patients.
The original MTS contain missing data, corresponding to measurements not collected for a given patient.
%, which are replaced with mean-imputation.
Data are in random order and the first $80\%$ are used as training set and the rest as test set.

The second medical dataset is \textit{Physionet}, which contains time series of peak-to-peak and RR intervals extracted from ECGs in the 2017 Atrial Fibrillation challenge~\cite{clifford2017af}. 
The MTS are divided into 4 classes: normal (N), atrial fibrillation (A), other symptoms (O) and noisy records ($\sim$).

% ================ MISS VAL ================
\subsection{Quantitative evaluations of MTS representations in the presence of missing data}
\label{sec:missing_data}

\begin{figure}[th!]
	\centering
	\subfigure[TCK ($\mathbf{K}$)]{
	\includegraphics[keepaspectratio,width=0.3\columnwidth]{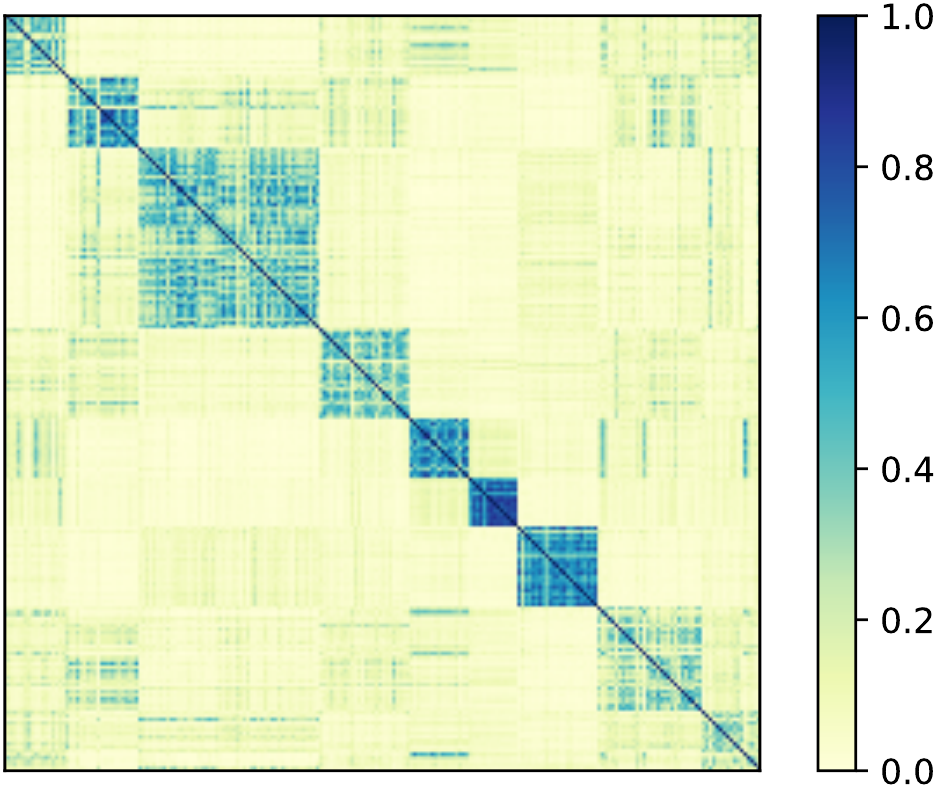}
	}
	%~
	%\subfigure[TAE]{
	%\includegraphics[keepaspectratio,width=0.45\columnwidth,trim={3.5cm 1cm 2cm 1cm},clip]{Code_prod_TAE.pdf}
	%}
	~
	\subfigure[TKAE ($\mathbf{ZZ}^T$)]{
	\includegraphics[keepaspectratio,width=0.3\columnwidth]{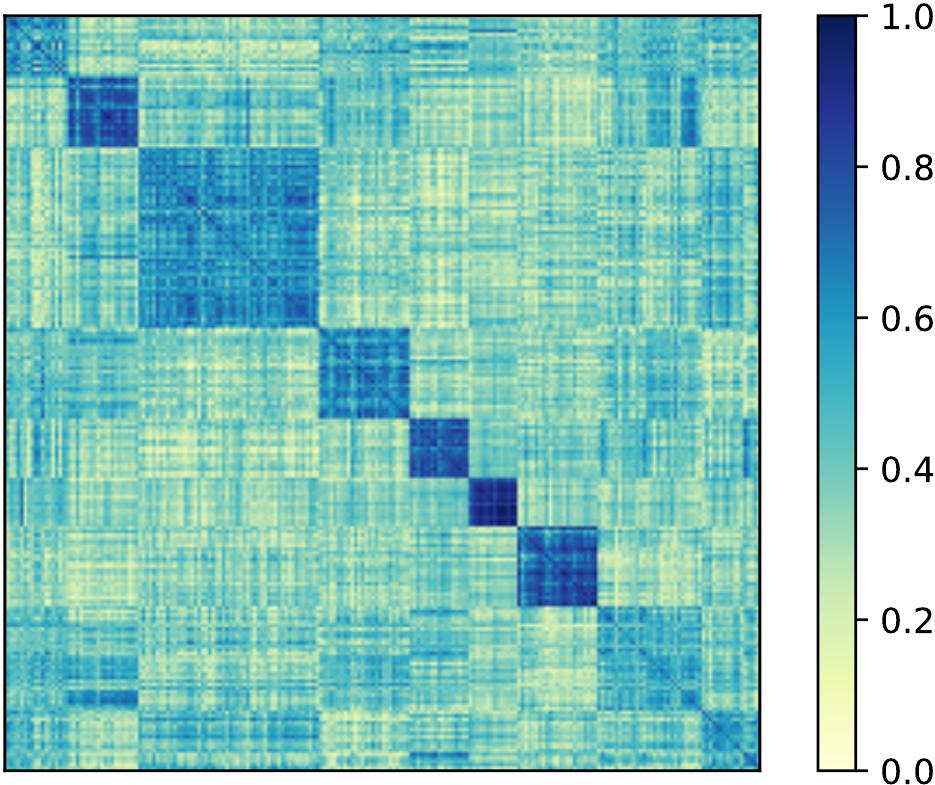}	
	}
\caption{Test set of Jp. Vow. with $80\%$ of missing data. (a) prior $\mathbf{K}$ computed with TCK in input space; 
(b) dot products $\mathbf{Z}\mathbf{Z}^T$ of the representations in TKAE.}
	\label{fig:alignment}
\end{figure}

% Jp vowel experiment
\paragraph{\textbf{Controlled experiments and sensitivity analysis}}
To evaluate the effect of kernel alignment when the MTS contain missing data, we perform a controlled experiment where we compare the representations learned by the TAE ($\alpha = 0$) and the TKAE ($\alpha \neq 0$) on the \textit{Jp. Vow.} dataset.
This dataset does not originally contain missing data. However, similarly to previous studies~\cite{Baydogan2016, mikalsen2017time}, we inject missing data in a controlled way by randomly removing a certain percentage of values.
We vary such percentage from $10\%$ to $90\%$, evaluating each time the reconstruction MSE and classification accuracy of the TAE and the TKAE encodings using $k$NN with $k=3$.
We apply zero imputation to replace missing data in the MTS. % fed both to TAE and TKAE, which is equivalent to mean imputation as data are standardized.
The TAE and the TKAE are configured with 2 LSTM cells, $p_s=0.9$ and $\lambda=0.001$. In the TKAE, $\alpha = 0.1$.
In Fig.~\ref{fig:alignment}, we show the kernel matrix $\mathbf{K}$ yielded by the TCK and the dot products in $\mathbf{Z}\mathbf{Z}^T$ of the representations of the test set when $80\%$ of the data are missing.
$\mathbf{Z}\mathbf{Z}^T$ is very similar to the TCK matrix, as they are both characterized by a block structure indicating that intra-class similarities in the 9 classes are much higher than inter-class similarities.

\begin{figure}[ht!]
	\centering
    \subfigure[Varying missingness]{
        \includegraphics[width=0.3\columnwidth]{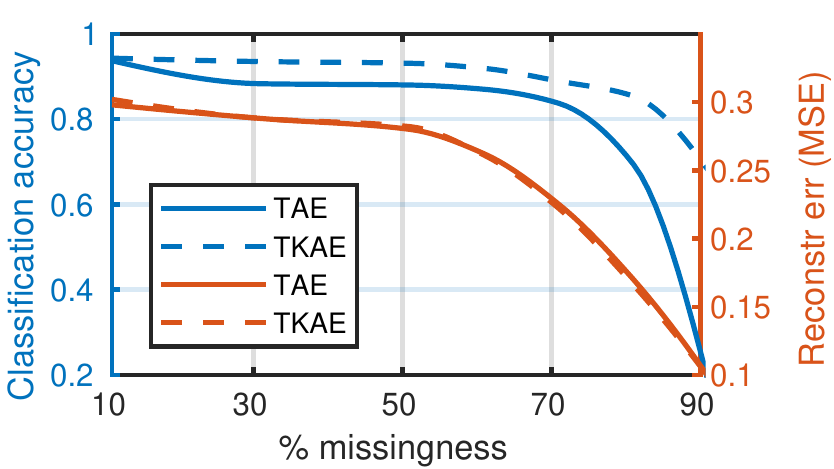}
    }\hspace{-0.5em}
	~
	\subfigure[Varying $\alpha$]{
        \includegraphics[width=0.3\columnwidth]{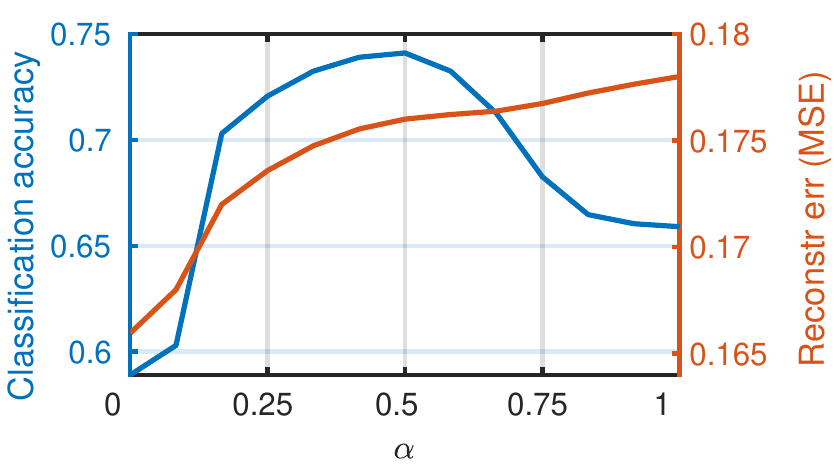}
	}\hspace{-0.5em}
    ~
	\subfigure[Varying $\lambda$]{
        \includegraphics[width=0.3\columnwidth]{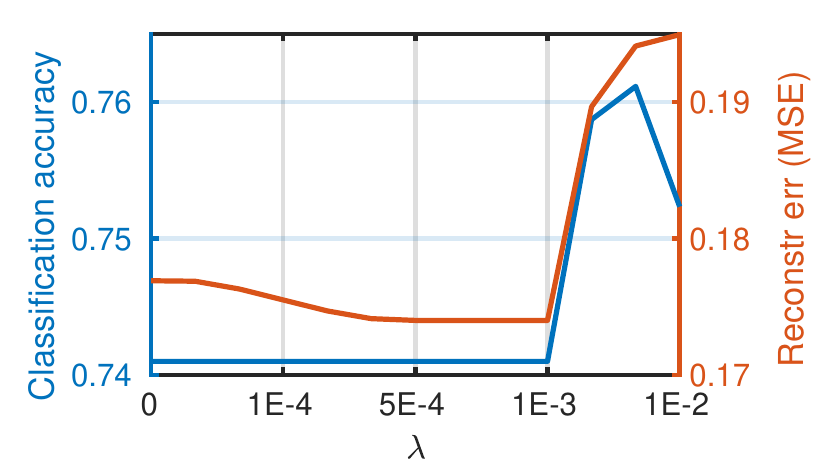}
	}

    \caption{\hspace{0pt}Classification accuracy (in blue) and reconstruction MSE (in red) on Japanese Vowels dataset. In (a), TAE and TKAE results are reported as a function of the missing values percentage. Panel (b) reports the sensitivity analysis for the parameter $\alpha$, when $\lambda=0$ is fixed and the percentage of missing values is $80\%$. Panel (c) reports the sensitivity analysis for the parameter $\lambda$, when $\alpha=0.5$ is fixed and the percentage of missing values is $80\%$.}
	\label{fig:miss}
\end{figure}

Fig.~\ref{fig:miss}(a) shows how the classification accuracy and reconstruction error of the TAE and the TKAE vary as we increase the amount of missing data.
The classification accuracy (blue lines) does not decrease in the TKAE when the data contain up to 50\% missing values and is always higher than in the TAE.
When $90\%$ of the data are missing, the TKAE still achieves a classification accuracy of 0.7, while for the TAE it drops to 0.1.
We note that the reconstruction MSE decreases for a higher amount of missingness. The reason is that by using imputation, a higher amount of missingness introduces more constant values and, therefore, there is less information to compress.

Fig.~\ref{fig:miss}(b) reports the classification accuracy and reconstruction error, by varying the kernel alignment parameter $\alpha$ when $\lambda=0$ is fixed and the percentage of missing values is $80\%$.
It is possible to note that there is quite a large interval ranging from 0.25 to 0.7 where the classification accuracy is above 70\%.
We also observe that the alignment term does not compromise input reconstruction, since the MSE does not significantly change as $\alpha$ is increased.

Finally, Fig.~\ref{fig:miss}(c) reports the classification accuracy and reconstruction error, by varying the $L_2$ regularization parameter $\lambda$,  when $\alpha=0.5$ is fixed and the percentage of missing values is $80\%$. 
The small changes in the results demonstrate that the TKAE is not very sensitive to this hyperparameter.

% CLASSIFICATION
\paragraph{\textbf{Classification of MTS representations in the presence of missing data}} In this second experiment, we analyze the MTS from the EHR, Japanese Vowels, and Arabic Digits datasets.
Since the last two datasets are complete, missingness is introduced by removing 80\% of the samples.
The performance is assessed by classifying the representations of the test set generated by each model.
We include in the comparison the representations generated by PCA, a standard AE, the TAE, and the Encoder-Decoder scheme for Anomaly Detection (EncDec-AD), which is a state-of-the-art architecture that generates vectorial representations of MTS~\cite{malhotra2016lstm}. The main differences between EncDec-AD and the TAE, are the lack of the bidirectional encoder in EncDec-AD and a deep architecture in TAE obtained by stacking multiple RNN layers in both the encoder and the decoder.
The networks considered for this experiment are configured with the parameters reported in Tab.~\ref{tab:class_config}.

% :::::::::::::::::::::::::::: TAB CLASS CONFIG ::::::::::::::::::::::::::::
\bgroup
\def\arraystretch{0.9} %vertical padding
\setlength\tabcolsep{.5em} %horizontal padding
\begin{table*}[!ht]
\footnotesize
\centering
\caption{Optimal hyperparameters found with cross-validation. For AE: type of activation function ($\psi$), $l_2$ regularization ($\lambda$), and tied weights in the decoder (tw). For TAE and TKAE: type of cell $\times$ number of layers, probability of scheduled sampling ($p_s$), kernel alignment ($\alpha$), and $l_2$ regularization ($\lambda$).}
\label{tab:class_config}
\begin{tabular}{lc|ccc|c|cc|c}
\cmidrule[1.5pt]{1-9}
\multirow{ 2}{*}{\textbf{Dataset}} & \multirow{ 2}{*}{$\boldsymbol{D_z}$} & \multicolumn{3}{c|}{\textbf{AE}} & \multicolumn{1}{c|}{\textbf{EncDec-AD, TAE, TKAE}} & \multicolumn{2}{c|}{\textbf{TAE, TKAE}} & \textbf{TKAE} \\
& & $\boldsymbol{\psi}$ & $\boldsymbol{\lambda}$ & \textbf{tw} & $\boldsymbol{\lambda}$ & \textbf{cell} & $\boldsymbol{p_s}$ & $\boldsymbol{\alpha}$ \\
\cmidrule[.5pt]{1-9}
EHR         & 10 & lin. & 0.001 & yes & 0     & GRU$\times$2  & 0.9 & 0.1    \\ 
Jp. Vow.    & 10 & lin. & 0.001 & no  & 0.001 & LSTM$\times$2 & 0.8 & 0.1   \\ 
Arab. Dig.  & 10 & lin. & 0     & yes & 0     & LSTM$\times$2 & 1.0 & 0.95  \\
\cmidrule[1.5pt]{1-9}
\end{tabular}
\end{table*}
\egroup
% :::::::::::::::::::::::::::::::::::::::::::::::::::::::::::::::::::::

Beside classification accuracy, in Tab.~\ref{tab:blood} we also report the F1 score that handles class imbalance.
For example, in the EHR data the \textit{infected} class is under-represented compared to \textit{not infected}.

% :::::::::::::::::::::::::: TAB BLOOD ::::::::::::::::::::::::::
\bgroup
\def\arraystretch{0.95} %vertical padding
\setlength\tabcolsep{.4em} %horizontal padding
\begin{table}[!ht]
\small
\centering
\caption{Classification of the blood data. F1 score is calculated considering \textit{infected} as ``positive'' class.}
\label{tab:blood}
\begin{tabular}{lcccccc}
\cmidrule[1.5pt]{1-7}
 & \multicolumn{2}{c}{\textbf{EHR}} & \multicolumn{2}{c}{\textbf{Jp. Vow.}} & \multicolumn{2}{c}{\textbf{Arab. Digits}}\\
%\cmidrule[.5pt]{1-7}
\textbf{Method} & Accuracy & F1 score & Accuracy & F1 score & Accuracy & F1 score \\
\cmidrule[.5pt]{1-7}
PCA         & 83.5 & 65.1 & 76.8 & 76.7 & 84.8 & 84.8 \\
AE          & 84.6$\pm$0.13 & 67.5$\pm$0.34 & 78.1$\pm$0.02 & 78.2$\pm$0.03 & 85.1$\pm$0.01 & 85.1$\pm$0.01 \\ 
% KAE & 0.869 & 0.741 \\
EncDec-AD   & 82.9$\pm$0.07 & 60.5$\pm$0.18 & 75.9$\pm$0.05 & 76.4$\pm$0.05 & 66.3$\pm$0.14 & 65.1$\pm$0.11 \\
TAE         & 85.3$\pm$0.021 & 68.2$\pm$0.22 & 78.6$\pm$0.17 & 79.1$\pm$0.15 & 73.1$\pm$0.09 & 72.8$\pm$0.11\\	
TKAE        & \textbf{89.9}$\pm$0.22 & \textbf{80.2}$\pm$0.47 & \textbf{82.4}$\pm$0.01 & \textbf{82.6}$\pm$0.01 & \textbf{86.8}$\pm$0.06 & \textbf{86.7}$\pm$0.02 \\
\cmidrule[1.5pt]{1-7}
\end{tabular}
\end{table}
\egroup
% :::::::::::::::::::::::::::::::::::::::::::::::::::::::::::::::

\begin{figure}[ht!]
	\centering
	\subfigure[TAE]{
	\includegraphics[keepaspectratio,width=0.35\columnwidth]{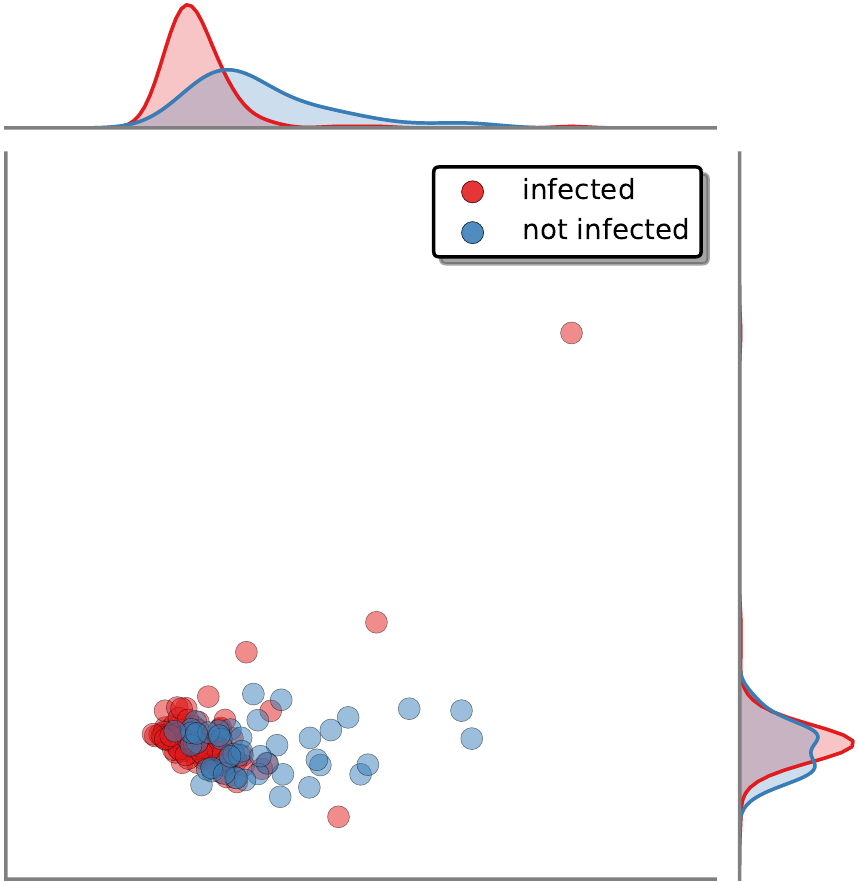}
	}
	~
	\subfigure[TKAE]{
	\includegraphics[keepaspectratio,width=0.35\columnwidth]{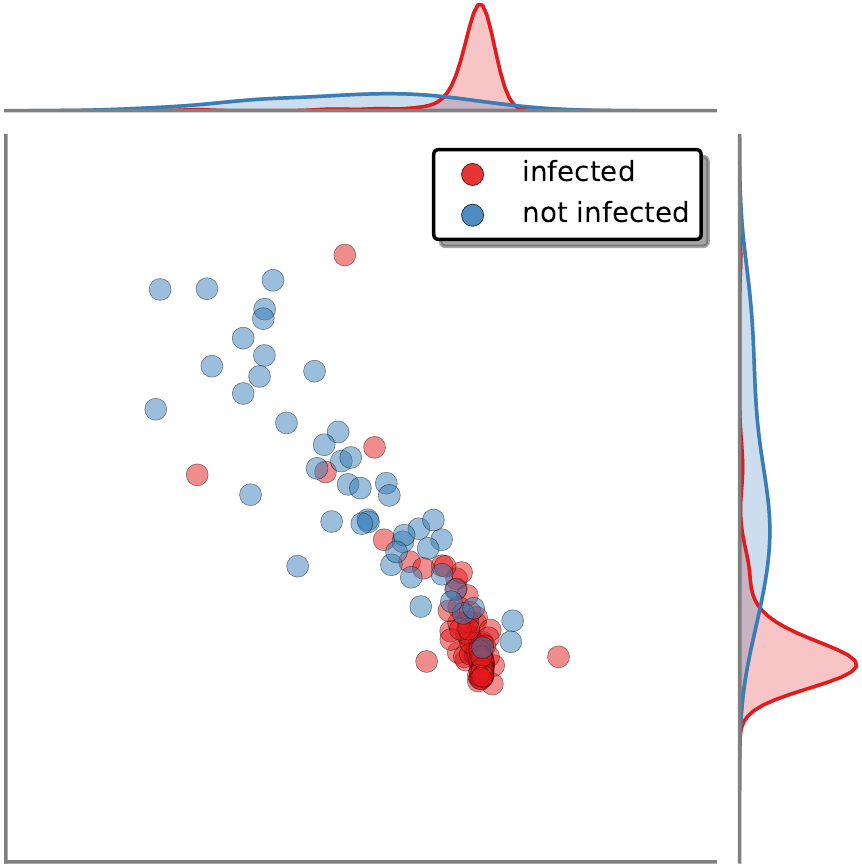}
	}
\caption{PCA and density of the two principal components of the representations yielded by TAE and TKAE on EHR dataset. The densities are computed with a kernel density estimator.}
	\label{fig:pca_blood}
\end{figure}

As expected, the AE achieves consistently better results than PCA.
In all the experiments, we observe that the TAE performs better than the EncDec-AD, which indicates the importance of the bidirectional encoder to learn MTS representations.
Except for the Arabic Digits dataset, the TAE performs also slightly better than the AE. 
However, when using kernel alignment the performance is boosted, as indicated by the results of the TKAE that are always the best in each task.
In particular, we observe that the TKAE representations achieve the best accuracy and a much higher F1 score in the EHR dataset.
Fig.~\ref{fig:pca_blood} depicts the first two principal components of the representations from the EHR dataset learned by the TAE and by the TKAE.
It is possible to recognize the effect of the kernel alignment, as the densities of the components relative to different classes become more separated.

% ================ Case Studies ================
\subsection{Decoder-based frameworks}
\label{sec:case_stud}

% ----- Imputation ------
\paragraph{\textbf{Imputation of missing data}} 
To perform missing data imputation, we modify the loss function of the TKAE slightly and exploit the decoder to impute missing values in the MTS.
In the presence of missing data, the reconstruction MSE of the loss function can be modified to account only for non-imputed values,
\begin{equation}
\label{eq:rec_miss}
L_r = - \sum \limits_{t} \left( (\mathbf{x}_t -\tilde{\mathbf{x}}_t)m_t \right)^2 / \sum \limits_{t} m_t,
\end{equation}
where $m_t=0$ if $\mathbf{x}_t$ is imputed and 1 otherwise.
In this way, the decoder is not constrained to reproduce the values that are imputed and, instead, freely assigns values to the entries that are originally missing. % this limit the biases introduced by the imputation.
Thus, we can exploit the generalization capability of the decoder to provide an alternative form of imputation, which depends on the nonlinear relationships existing in the training data.
A similar principle is followed by denoising AEs (DAEs)~\cite{vincent2010stacked}, as they try to reconstruct the original input from a corrupted version of it, where some entries are randomly removed.
Therefore, after training, DAEs can be exploited to impute missing values on new unseen data~\cite{beaulieu2016semi, gondara2017multiple}.

We randomly remove approximately $50\%$ of the values from 5 of the datasets described in Tab.~\ref{tab:dataset_details} and compare the capability of the TKAE to reconstruct missing values with respect to other imputation techniques.
As baselines, we consider mean imputation, last occurrence carried forward (LOCF), and DAE imputation~\cite{beaulieu2017missing}.
For the TKAE and the DAE, we use the optimal configurations identified with cross-validation, reported in Tab.~\ref{tab:imp_config}.
In the TKAE we replace the $L_r$ term with \eqref{eq:rec_miss} and we set $\alpha=0.1$.
In the DAE, we apply a stochastic corruption of the inputs, by randomly setting input values to $0$ with probability $0.5$.

% :::::::::::::::::::::::::::: TAB IMP CONFIG ::::::::::::::::::::::::::::
\bgroup
\def\arraystretch{0.9} %vertical padding
\setlength\tabcolsep{.3em} %horizontal padding
\begin{table*}[!ht]
\footnotesize
\centering
\caption{Optimal hyperparameters found with cross-validation. For DAE: type of activation function ($\psi$), $\lambda$ of $l_2$ regularization, and tied weights in the decoder (tw). For TKAE: type of cell, number of layers, probability of scheduled sampling ($p_s$), and $\lambda$ of $l_2$ regularization.}
\label{tab:imp_config}
\begin{tabular}{lc|ccc|ccc}
\cmidrule[1.5pt]{1-8}
\multirow{ 2}{*}{\textbf{Dataset}} & \multirow{ 2}{*}{$\boldsymbol{D_z}$} & \multicolumn{3}{c|}{\textbf{DAE}} & \multicolumn{3}{c}{\textbf{TKAE}} \\
& & $\boldsymbol{\psi}$ &  $\boldsymbol{\lambda}$ & \textbf{tw} & \textbf{cell} & $\boldsymbol{p_s}$ & $\boldsymbol{\lambda}$ \\
\cmidrule[.5pt]{1-8}
ECG         & 10 & lin. & 0     & no  & GRU$\times$1  & 1.0 & 0     \\ 
Lbras       & 5  & sig. & 0.001 & no  & LSTM$\times$2 & 0.9 & 0.001 \\ 
Wafer       & 10 & lin. & 0     & no  & LSTM$\times$2 & 1.0 & 0     \\ 
Jp. Vow.    & 10 & lin. & 0.001 & no  & LSTM$\times$2 & 0.8 & 0.001 \\ 
Auslan      & 10 & lin. & 0     & yes & LSTM$\times$2 & 1.0 & 0     \\ 
\cmidrule[1.5pt]{1-8}
\end{tabular}
\end{table*}
\egroup
% :::::::::::::::::::::::::::::::::::::::::::::::::::::::::::::::::::::

% ::::::::::::::::::::::::::: TAB IMP ::::::::::::::::::::::::::::
\bgroup
\def\arraystretch{0.95} %vertical padding
\setlength\tabcolsep{.3em} %horizontal padding
\begin{table*}[!htp]
\small
\centering
\caption{MSE and Pearson correlation (CORR) of the MTS where missing values are imputed using different methods, with respect to the original MTS (without missing values). Best and second best results are highlighted in dark and light blue, respectively.}
\label{tab:imputation}
\begin{tabular}{l|cc|cc|cc|cc}
\cmidrule[1.5pt]{1-9}
\multirow{2}{*}{\textbf{Dataset}} & \multicolumn{2}{c|}{\textbf{Mean Imp.}} & \multicolumn{2}{c|}{\textbf{LOCF}} & \multicolumn{2}{c|}{\textbf{DAE}} & \multicolumn{2}{c}{\textbf{TKAE}} \\
& \textbf{MSE} & \textbf{CORR} & \textbf{MSE} & \textbf{CORR} & \textbf{MSE} & \textbf{CORR} & \textbf{MSE} & \textbf{CORR} \\
\cmidrule[.5pt]{1-9}
ECG   & 0.883 & 0.702 & 0.393 & 0.884 & \cellcolor{blue!15} 0.157$\pm$0.004 & \cellcolor{blue!15} 0.956$\pm$0.001 & \cellcolor{blue!35} \textbf{0.151}$\pm$0.003 & \cellcolor{blue!35} \textbf{0.956}$\pm$0.001 \\
Libras & 0.505 & 0.666 & 0.085 & 0.949 & \cellcolor{blue!15} 0.050$\pm$0.001 & \cellcolor{blue!15} 0.970$\pm$0.001 & \cellcolor{blue!35} \textbf{0.029}$\pm$0.002 & \cellcolor{blue!35} \textbf{0.978}$\pm$0.002 \\
Wafer  & 0.561 & 0.695 & 0.226 & 0.911 & \cellcolor{blue!15} 0.199$\pm$0.017 & \cellcolor{blue!15} 0.935$\pm$0.004 & \cellcolor{blue!35} \textbf{0.093}$\pm$0.007 & \cellcolor{blue!35} \textbf{0.964}$\pm$0.003 \\
Jp. Vow. & 0.502 & 0.699 & \cellcolor{blue!35} \textbf{0.084} & \cellcolor{blue!35} \textbf{0.954} & 0.132$\pm$0.001 & 0.926$\pm$0.000 & \cellcolor{blue!15} 0.114$\pm$0.003 & \cellcolor{blue!15} 0.938$\pm$0.001 \\
Auslan & 0.532 & 0.613 & 0.379 & 0.746 & \cellcolor{blue!15} 0.145$\pm$0.002 & \cellcolor{blue!15} 0.873$\pm$0.005 & \cellcolor{blue!35} \textbf{0.087}$\pm$0.001 & \cellcolor{blue!35} \textbf{0.941}$\pm$0.002 \\
\cmidrule[1.5pt]{1-9}
\end{tabular}
\end{table*}
\egroup
% :::::::::::::::::::::::::::::::::::::::::::::::::::::::::::::::

In Tab.~\ref{tab:imputation} we report the MSE and the Pearson correlation (CORR) of the MTS with imputed missing values, with respect to the original MTS.
We observe that in 4 of the 5 datasets the TKAE yields the most accurate reconstruction of the true input, followed by the DAE.
However, in \textit{Jp. Vow.} LOCF imputation allows retrieval of missing values with the highest accuracy.
This can be explained by the fact that in the MTS of the \textit{Jp. Vow.} dataset very similar values are repeated for several time intervals. 
However, we notice that also in this case the TKAE achieves the second-best result and it outperforms the DAE.

% ---- One-class classification -----
\paragraph{\textbf{One-class classification}} One-class classification and anomaly detection are applied in several domains, including healthcare~\cite{CARRERA2019482}, where non-nominal samples are scarce and often unavailable during training~\cite{irigoien2014towards}.
When dealing with MTS data, RNN-based approaches have been adopted to perform anomaly detection tasks~~\cite{malhotra2016lstm, zhang2018deep}.
The methods based on dimensionality reduction procedures, such as AEs and energy based models~\cite{pmlr-v48-zhai16, CHAKRABORTY2019161} rely on the assumption that anomalous samples do not belong to the subspace containing nominal data, which is learned during training.
Therefore, the representations generated by the trained model for samples of a new, unseen class will arguably fail to capture important characteristics.
Consequently, for those samples an AE would yield large reconstruction errors, which we consider as the classification scores for the new class.

% :::::::::::::::::::::::::::: TAB AF :::::::::::::::::::::::::::
\bgroup
\def\arraystretch{0.95} %vertical padding
\setlength\tabcolsep{.2em} %horizontal padding
\begin{SCtable}[2][!ht]
\small
\centering
\caption{AUC obtained by different one-class classification methods in detecting the MTS of atrial fibrillation class, which is not present in the training set.}
\label{fig:anomaly_detect}
\begin{tabular}{lc}
\cmidrule[1.5pt]{1-2}
\textbf{Method} & \textbf{AUC} \\
\cmidrule[.5pt]{1-2}
OCSVM  &  0.713 \\ 
IF     &  0.662$\pm$0.01 \\
PCA    &  0.707 \\
AE     &  0.712$\pm$0.001 \\ 
EncDec-AD    &  0.719$\pm$0.007 \\	
TAE    &  0.728$\pm$0.005 \\
TKAE   &  \textbf{0.732}$\pm$0.006 \\
\cmidrule[1.5pt]{1-2}
\end{tabular}
\end{SCtable}
\egroup
% :::::::::::::::::::::::::::::::::::::::::::::::::::::::::::::::

For this task, we consider the real-world data from the \textit{Physionet} dataset.
By following a commonly adopted procedure~\cite{doi:10.3102/00346543074004525}, we simulate missing data by randomly removing approximately $50\%$ of the entries in each MTS and then we exclude samples of class A from the training set (which are then considered as non-nominal).
We evaluated the performance of the TKAE, the AE, and PCA in detecting class A in a test set containing samples of all classes (N,A,O,$\sim$).
As performance measure, we considered the area under ROC curve (AUC) and compared the performance also with two baseline classifiers: one-class SVM (OCSVM)~\cite{GUERBAI2015103} and Isolation Forests (IF)~\cite{DOMINGUES2018406}. 
The optimal configurations are: $D_z = 10$; EncDec-AD, TAE and TKAE with $\lambda=0$; TAE and TKAE with 1 layer of GRU cells, $p_s=0.9$; TKAE with $\alpha=0.2$; AE with non-linear decoder, no tied weights, and $\lambda=0$; OCSVM with rbf kernel width $\gamma = 0.7$ and $\nu = 0.5$; IF with contamination $0.5$.
Results in Tab.~\ref{fig:anomaly_detect} show that the TKAE scores the highest AUC.

%\begin{figure}[ht!]
%	\centering
%	\includegraphics[keepaspectratio,width=0.9\columnwidth]{Anomaly_detect.pdf}	
%    \caption{Anomaly detection based on MSE thresholding. \commentF{change or remove if not enough space.}}
%	\label{fig:anomaly_detect}
%\end{figure}

%%%%%%%%%%%%%%%%%%%%%%%%%%%%%%%%%%
%%%%%%%%%%%% ANALYSIS %%%%%%%%%%%%
%%%%%%%%%%%%%%%%%%%%%%%%%%%%%%%%%%
\section{Comparative analysis of recurrent and feed-forward architectures for learning compressed representations of MTS}
\label{sec:analysis}

So far, we demonstrated the capability of the TKAE to learn \textit{good} representations even in the presence of missing data, thanks to the kernel alignment with the TCK.
Another characterizing component of the TKAE is the use of RNNs in both encoder and decoder.
Recurrent layers have been successfully applied in \textit{seq2seq} models~\cite{sutskever2014sequence} to encode different types of sequential data, such as text and videos.
However, their application to real-valued MTS has been limited so far and it is not clear yet in which cases recurrent AEs work well.
Therefore, in this section we investigate when RNNs can represent MTS better than a feed-forward architecture, which processes the whole MTS at once using padding to deal with inputs of variable lengths.
We show that in most cases an AE with RNNs encodes well the MTS, which justifies our choice in the design of the TKAE.
However, we also report examples of negative results where RNNs fail to process MTS exhibiting certain properties.

Since we want to focus only on the effects of using recurrent layers in generating compressed representations, in the following we do not use kernel alignment, but we consider only the TAE (i.e., TKAE with $\alpha=0$). Synthetic data are generated to study specific MTS properties in controlled environments.

% -------------- SIN DIFFERENT FREQS ---------------
\paragraph{\textbf{Time series with different frequencies}}

Here, we evaluate the capability of the TAE to compress periodic signals having different frequencies and phases.
We generate a dataset of sinusoids $y(t) = \text{sin}(a \cdot t+b)$, where $a, b$ are drawn from $\mathcal{N}(0,1)$ and $t \in [0,100]$.
The proposed task is closely related to the multiple superimposed oscillators, studied in pattern generation and frequency modulation~\cite{sussillo2009generating}.
The training and test sets contain $200$ and $1000$ samples, respectively.
We let $D_z = 5$ and the optimal configurations are: AE with nonlinear decoder and $\lambda=0.001$; TAE with 2 layers of LSTM cells, $\lambda=0$, and $p_s=1.0$.
The reconstruction MSE on the test set is $0.41$ for PCA, $0.212$ for the AE, and $0.013$ for the TAE.

\begin{SCfigure}[0.8][th!]
    \centering
    \includegraphics[keepaspectratio,width=0.5\columnwidth]{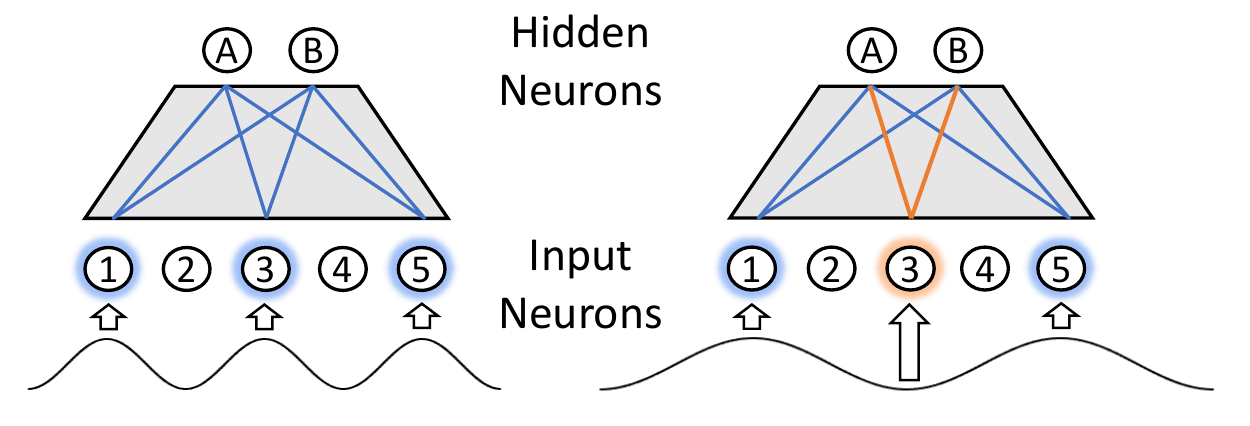}
    \caption{Periodic inputs with different frequencies generate different activation patterns in AEs. 
    It is not possible to learn connections weights that preserve neurons couplings for each frequency.}
    \label{fig:AEsins}
\end{SCfigure}

Both PCA and the AE process the entire time series at once. This may appear an advantage with respect to the TAE, which stores information in memory for the whole sequence length before yielding the final representation.
Nonetheless, the TAE produces a better reconstruction, while the AE (and PCA) is unsuitable for this task.
Indeed, in AEs a given time step $t$ in each MTS is always processed by the same input neuron.
For periodic signals, the training procedure tries to couple neurons associated to time steps with the same phase, by assigning similar weights to their connections. However, these couplings always change if inputs have different frequencies (see Fig.\ref{fig:AEsins}).
Therefore, training in the AE never converges as it is impossible to learn a model that generalizes well for each frequency.
On the other hand, thanks to its recurrent architecture, the TAE can naturally handle inputs of different frequencies as there is no pairing between structural parameters and time steps.

\begin{SCfigure}[0.8][th!]
    \centering
    \includegraphics[keepaspectratio,width=0.35\textwidth]{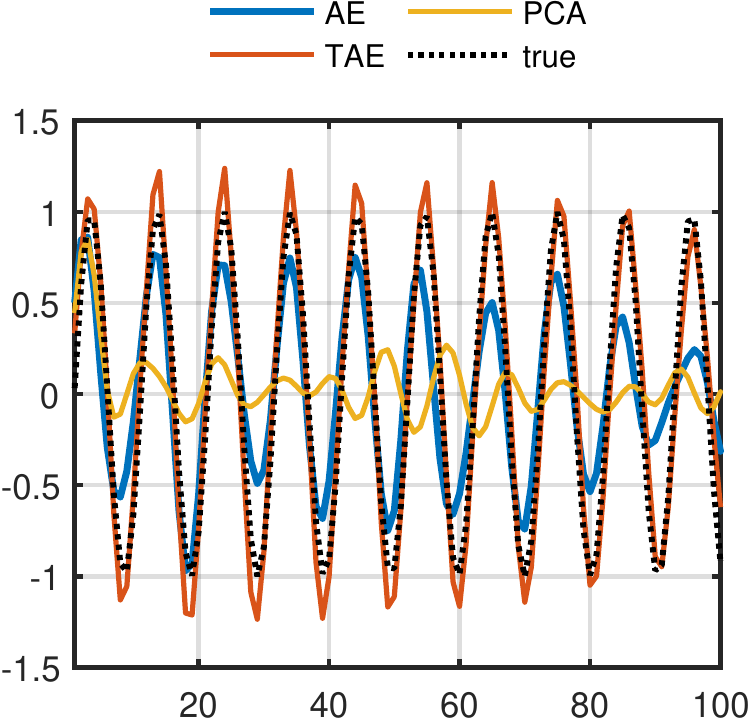}	
    \caption{\hspace{0pt}Reconstructions obtained by PCA, AE, and TAE on a sample sinusoid, whose frequency and phase are randomly chosen.}
    \label{fig:sin_test}
\end{SCfigure}

Fig.~\ref{fig:sin_test} shows the reconstruction of one sample time series.
The lower quality of the reconstruction yielded by the AE and by PCA can be immediately noticed. 
Additionally, since they are unable to reproduce the dynamics of each sample, they rather adopt a more conservative behavior and output signals with lower amplitudes that are closer (in a mean square sense) to the ``average'' of all the random sinusoids in the dataset.

% -------------- TS DIFFERENT LEN ---------------
\paragraph{\textbf{Time series with variable lengths}}

While the TAE can process MTS of different length, the standard AE and PCA require inputs of fixed size. 
The common workaround also followed in this work, is to pad the shorter MTS with zeros~\cite{mann2008smoothing}.
To systematically study the performance of the different methods when the MTS have fixed or variable length, we generate data by integrating the following system of first-order Ordinary Differential Equations (ODE):
\begin{equation}
\label{eq:ode}
\frac{d\mathbf{y}}{dt} = \mathbf{A}\text{tanh}\left(\mathbf{y}(t)\right),
\end{equation}
where $\mathbf{y} \in \mathbb{R}^{V}$, $\mathbf{A} \in \mathbb{R}^{V \times V}$ is a matrix with 50\% sparsity and elements uniformly drawn in $[-0.5,0.5]$.
To guarantee system stability~\cite{bianchi2017multiplex}, we set the spectral radius of $\mathbf{A}$ to $0.8$.
$\tanh(\cdot)$ is applied component-wise and introduces nonlinear dependencies among the variables.
A MTS $\mathbf{x} \in \mathbb{R}^{T \times V}$ is obtained by integrating \eqref{eq:ode} for $T$ steps, starting from a random initial condition $\mathbf{y}(0)$.
Since a generic deterministic dynamical system can be described by an ODE system, these synthetic MTS can represent many real data.

We generate two different datasets of MTS with $V=10$ variables, each one with 400 and 1000 samples for training and test set, respectively.
The first, \texttt{ODEfix}, contains MTS with same length $T=90$, while in the second, \texttt{ODEvar}, each MTS has a random length $T \in [30,90]$.
We let $D_z = 10$ and compare the reconstruction MSE of PCA, the AE, and the TAE.
The optimal configurations for this task are: AE with $\lambda = 0.001$ and linear decoder; TAE with 1 LSTM layer, $\lambda = 0.001$, and $p_s=0.9$.
The average results obtained for $10$ independent random generations of the data ($\mathbf{A}$) and initialization of AE and TAE are reported in Tab.~\ref{tab:ODE1}.
\begin{table}[!ht]
\footnotesize
\centering
\caption{Average reconstruction MSE of MTS with fixed (\texttt{ODEfix}) and variable (\texttt{ODEvar}) length.}
\label{tab:ODE1}
\begin{tabular}{l|ccc}
\cmidrule[1.5pt]{1-4}
\multirow{1}{*}{\textbf{Dataset}} & \multicolumn{1}{c}{\textbf{PCA}} & \multicolumn{1}{c}{\textbf{AE}} & \multicolumn{1}{c}{\textbf{TAE}} \\
\cmidrule[.5pt]{1-4}
\texttt{ODEfix} & 0.018 & \textbf{0.004} & 0.060 \\
\texttt{ODEvar} & 0.718 & 0.676 & \textbf{0.185} \\
\cmidrule[1.5pt]{1-4}
\end{tabular}
\end{table}

In \texttt{ODEfix}, both the AE and PCA yield almost perfect reconstructions, which is expected due to the simplicity of the task.
However, they perform worse in \texttt{ODEvar} despite the presence of many padded values and a consequent lower amount (on average) of information to encode in the compressed representation.
On the other hand, the TAE naturally deals with variable-length inputs, since once the input sequence terminates its state and model weights during the training are no longer updated.

% -------------- MULTIVARIATE TEST ---------------
\paragraph{\textbf{Dealing with a large number of variates and time steps}}

% increasing V
To test the ability to learn compressed representations when the number of variates in the MTS increases, starting from \eqref{eq:ode} we generate four datasets \texttt{ODE5}, \texttt{ODE10}, \texttt{ODE15}, and \texttt{ODE20}, obtained by setting $V = \{ 5, 10, 15, 20 \}$. 
The number of time steps is fixed to $T=50$ in each dataset.
We let $D_z=10$; TAE is configured with 2 layers of LSTM and $p_s = 0.9$; $\lambda$ is $0.001$ in both the AE and the TAE.
We also include in the comparison an AE with tied weights in the (nonlinear) decoder, which has fewer parameters.
Reconstruction errors are reported in Tab. \ref{tab:ODE}.
We notice that the AE performs well on MTS characterized by low dimensionality, but performance degrades when $V$ assumes larger values.
Since the AE processes MTS unrolled into a unidimensional vector, the input size grows quickly as $V$ increases (one additional variable increases the input size by $T$). 
Accordingly, the number of parameters in the first dense layer scales-up quickly, possibly leading to overfitting. 
We also notice that the tied weights regularization, despite halving the number of trainable parameters, degrades performance in each case, possibly because it hinders too much the flexibility of the model.
On the other hand, the TAE complexity changes slowly, as only one single neuron is added for an additional input dimension.
As a consequence, we conclude that the TAE is the best performing model when the MTS have a large number of variates.
%
% :::::::::::::::::::::::::::: TAB. ODE RES ::::::::::::::::::::::::::::
\bgroup
\def\arraystretch{0.95} %vertical padding
\setlength\tabcolsep{.35em} %horizontal padding
\begin{table}[!ht]
\footnotesize
\centering
\caption{Average reconstruction MSE on the ODE task for different values of $V$, obtained by TAE, AE, AE with tied weights (tw) and PCA. For AE and TAE we report the number of trainable parameters (\#par). Best results are in bold.}
\label{tab:ODE}
\begin{tabular}{l|cc|cc|cc|c}
\cmidrule[1.5pt]{1-8}
\multirow{ 2}{*}{\textbf{Dataset}} & \multicolumn{2}{c|}{\textbf{TAE}} & \multicolumn{2}{c|}{\textbf{AE}} & \multicolumn{2}{c|}{\textbf{AE (tw)}} & \textbf{PCA} \\
& \textbf{MSE} & \textbf{\#par} & \textbf{MSE} & \textbf{\#par} & \textbf{MSE} & \textbf{\#par} & \textbf{MSE} \\
\cmidrule[.5pt]{1-8}
\texttt{ODE5}  & 0.019 & 6130 & \textbf{0.04} & 31170  & 0.014 & 15870 & 0.007  \\
\texttt{ODE10} & 0.060 & 6780 & \textbf{0.04}  & 61670  & 0.071 & 31370 & 0.018  \\
\texttt{ODE15} & \textbf{0.072} & 7430 & 0.106 & 92170  & 0.153 & 46870 & 0.174  \\
\texttt{ODE20} & \textbf{0.089} & 8080 & 0.121 & 122670 & 0.181 & 62370 & 0.211  \\
\cmidrule[1.5pt]{1-8}
\end{tabular}
\end{table}
\egroup

\begin{SCfigure}[0.8][th!]
\centering
\includegraphics[keepaspectratio,width=0.4\columnwidth]{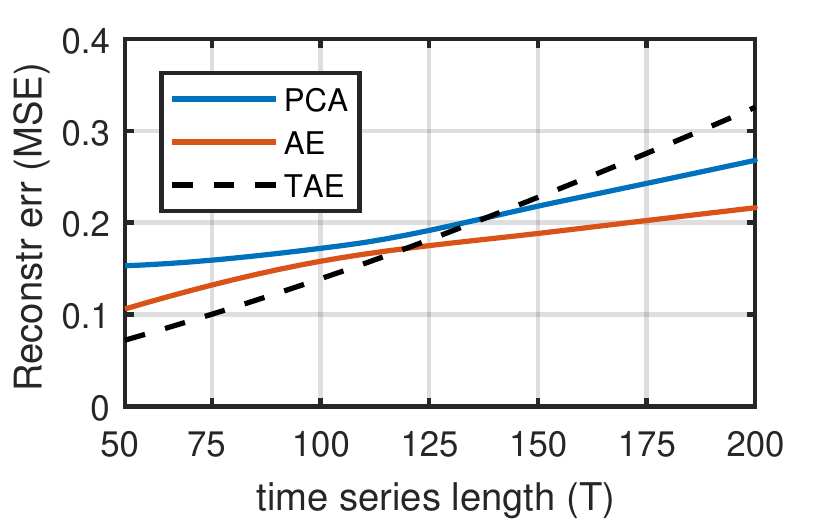}	
\caption{Reconstruction MSE when increasing length $T$ of MTS in \texttt{ODE15}. TAE performance decreases for large $T$.}
\label{fig:T_incr}
\end{SCfigure}

% increasing T
To study the performance as the lengths of MTS increase, we generate 8 datasets with the ODE system \eqref{eq:ode} by varying $T \in \{ 50,75,100,125,150,175,200 \}$, while keeping $V=15$ fixed.
In Fig. \ref{fig:T_incr}, we report the reconstruction errors and note that the TAE performance decays as $T$ increases.
RNNs excel in capturing recurring patterns in the input sequence and they can model extremely long sequences whenever they are characterized by a strong periodicity.
However, in this case there are no temporal patterns in the data that can be exploited by the RNNs to model the inputs. 
Therefore, the RNN dynamics do not converge to any fixed point and the modeling task becomes much more difficult as the input length increases.

%%%%%%%%%%%%%%%%%%%%%%%%%%%%%%%%%%%%%%%%%%%%%%%
%%%%%%%%%%%%%%%%% CONCLUSION %%%%%%%%%%%%%%%%%%
%%%%%%%%%%%%%%%%%%%%%%%%%%%%%%%%%%%%%%%%%%%%%%%
\section{Conclusion}
\label{sec:conclusion}

We proposed the temporal kernelized autoencoder, an RNN-based model for representing MTS with missing values as fixed-size vectors. 
Missing values in MTS are commonly found in domains such as healthcare and derive from measurement errors, incorrect data entry or lack of observations.
Through a kernel alignment performed with the time series cluster kernel, a similarity measure designed for MTS with missing data, our method learns compressed representations that preserve pairwise relationships defined in the original input space, even when data are heavily corrupted by missing values.\\
We showed that the representations learned by the TKAE can be exploited both in supervised and unsupervised tasks.
Experimental results, contrasted with other dimensionality reduction techniques on several datasets, showed that the TKAE representations are classified accurately also when the percentage of missing data is high.
Through sensitivity analysis, we showed that the kernel alignment has very little impact on the reconstruction error, demonstrating that the TKAE can learn \textit{good} representations even when using the alignment procedure.\\
After training, only the TKAE encoder is used to generate the representation of the MTS, while the decoder, which is learned as part of the optimization, remain unused.
To fully exploit the capabilities of the TKAE architecture, we considered two applications that take advantage of the decoder module. 
Specifically, we designed two frameworks based on dimensionality reduction and inverse mapping to the input space for imputing missing data and for one-class classification.
We showed that by thresholding the reconstruction error of the decoder, the TKAE is able to outperform competing approaches on these tasks.\\
We concluded our work by investigating which types of MTS are better modeled by a neural network auto-encoder with recurrent layers, rather than with feed-forward ones.
Our results showed that in most cases an RNN-based AE is the best architecture to generated good MTS representations. This motivated our design choice for the TKAE.
Our analysis revealed that an RNN excels in encoding short MTS with many variables, that are characterized by different lengths or by a varying periodicity.
However, when MTS are very long and do not contain temporal patterns that can be modeled by RNNs, better performance can be achieved by replacing recurrent layers in the TKAE with standard dense layers.

\section*{Acknowledgments}
This work was partially funded by the Norwegian Research Council FRIPRO grant no. 239844 on developing the \emph{Next Generation Learning Machines}.
The authors would like to thank Arthur Revhaug, Rolv-Ole Lindsetmo and Knut Magne Augestad, all clinicians currently or formerly affiliated with the gastrointestinal surgery department at the University Hospital of North Norway, for preparing the blood samples dataset.
LL gratefully acknowledges partial support of the Canada Research Chairs program.

\appendix

\section{Details of the TCK algorithm}
\label{app:tck}

A MTS $\mathbf{X} \in \mathbb{R}^{V \times T}$ is represented by a sequence of $V$ univariate time series (UTS) of length $T$,
$\mathbf{X} = \{ \mathbf{x}_v \in \mathbb{R}^T \: | \: v = 1,\dots,V\}$, being $V$ and $T$ the dimension and length of $\mathbf{X}$, respectively.
Given a dataset of $N$ samples, $\mathbf{X}^{(n)}$ denotes the $n$-th MTS and a binary MTS $\mathbf{R}^{(n)} \in \mathbb{R}^{V \times T}$ describes whether the realisation $x_v^{(n)}(t)$ in $\mathbf{X}$ is observed ($r_v^{(n)}(t) = 1$) or is missing ($r_v^{(n)}(t) = 0$).

\paragraph{\textbf{DiagGMM}}

The TCK kernel matrix is built by first fitting $G$ diagonal covariance GMM (DiagGMM) to the MTS dataset.
Each DiagGMM $g$ is parametrized by a time-dependent mean $\boldsymbol{\mu}_{gv} \in \mathbb{R}^T$ and a time-constant covariance matrix $\Sigma_g = diag\{\sigma_{g1}^2,...,\sigma_{gV}^2\}$, being $\sigma_{gv}^2$ the variance of UTS $v$.
Moreover, the data is assumed to be \textit{missing at random}, i.e. the missing elements are only dependent on the observed values.
Under these assumptions, missing data can be analytically integrated away~\cite{rubin1976inference} and the pdf for each incompletely observed MTS $\{\mathbf{X}, \mathbf{R}\}$ is given by
\begin{equation} 
\label{eq: p(x) gmm diag}
p(\mathbf{X} \: | \: \mathbf{R}, \: \Theta ) = \sum_{g=1}^G \theta_g \prod_{v=1}^V \prod_{t=1}^T  \mathcal{N} (x_v(t) \: | \: \boldsymbol{\mu}_{gv}(t), \sigma_{gv})^{r_v(t) }
\end{equation}

The conditional probabilities follows from Bayes' theorem,
\begin{equation} 
\label{eq: p(z|x) posterior}
\boldsymbol{\pi}_{g} %\equiv P(\mathbf{Y}_g = 1 \: | \: X, \: R, \: \Theta )  
= \frac{ \theta_g \prod_{v=1}^V \prod_{t=1}^T  \mathcal{N} \left(x_v(t) \: | \: \boldsymbol{\mu}_{gv}(t), \sigma_{gv}\right)^{r_v(t) }}{\sum_{g=1}^G \theta_g \prod_{v=1}^V \prod_{t=1}^T  \mathcal{N} \left(x_v(t) \: | \: \boldsymbol{\mu}_{gv}(t), \sigma_{gv}\right)^{r_v(t) }}.
\end{equation}
The parameters of the DiagGMM are trained by means of a maximum a posteriori expectation maximization algorithm, as described in \cite{mikalsen2017time}. 

\paragraph{\textbf{Ensemble generation}}

To ensure diversity in the ensemble, each GMM model has a different number of components from the interval $[2,C]$ and is trained $Q$ times, using random initial conditions and hyperparameters.
Specifically, $\mathcal{Q} = \{ q = (q_1,q_2) \: | \: q_1=1,\dots Q, \: q_2 = 2,\dots, C \} $ denotes the index set of the initial conditions and hyperparameters ($q_1$), and the number of components ($q_2$).
Moreover, each DiagGMM is trained on a subset of the original dataset, defined by a random set of the MTS samples, a random set $\mathcal{V}$ of $|\mathcal{V}| \leq V$ variables, and a randomly chosen time segment $\mathcal{T}, |\mathcal{T}| \leq T$. 
The inner products of the posterior distributions from each mixture component are then added up to build the final TCK kernel matrix.
Details are provided in Alg.~\ref{alg:algorithm}.

% :::::::::::::::::::::: ALGO TCK in-sample ::::::::::::::::::::::
\begin{algorithm}[t!]
%\footnotesize
\small
\caption{TCK kernel training}
\label{alg:algorithm}
\begin{algorithmic}[1]
\Require Training set of MTS $ \{ \mathbf{X}^{(n)}  \}_{n=1}^N$ , $Q$ initializations, $C$ maximal number of mixture components.
\State Initialize kernel matrix $\mathbf{K} = \boldsymbol{0}_{N \times N}  $.
\For{$q \in \mathcal{Q}$}
\State Compute posteriors $ \Pi^{(n)}(q) \equiv ( \pi_1^{(n)},\dots,\pi_{q_2}^{(n)} )^T $, by applying maximum a posteriori expectation maximization~\cite{mikalsen2017time} to the DiagGMM with $q_2$ components and by randomly selecting,
%{\setstretch{0.4}
\begin{itemize}
\item[i.] hyperparameters $\Omega(q) $,
\item[ii.] a time segment $ \mathcal{T}(q)  $ of length {\footnotesize $T_{min} \leq  |\mathcal{T}(q)| \: \leq \: T_{max}$ },
\item[iii.] attributes $\mathcal{V}(q)$, with cardinality {\footnotesize$V_{min} \leq |\mathcal{V}(q)| \leq V_{max}$},
\item[iv.] a subset of MTS, $\eta(q) $, with {\footnotesize$N_{min} \leq |\eta(q)| \leq N$},
\item[v.] initialization of the mixture parameters $ \Theta(q) $.
\end{itemize}
%}
\State Update kernel matrix, $\mathbf{K}_{nm} = \mathbf{K}_{nm} + \frac{\Pi^{(n)}(q)^T \Pi^{(m)}(q)}{ \| \Pi^{(n)}(q) \| \| \Pi^{(m)}(q) \| } $.
\EndFor
\Ensure TCK matrix $\mathbf{K}$.
\end{algorithmic}
\end{algorithm}
% :::::::::::::::::::::::::::::::::::::::::::::::::::::::::::::::::

\bibliographystyle{abbrv}
\bibliography{Biblio}

\end{document}